# Automatic Assignment of Radiology Examination Protocols Using Pre-trained Language Models with Knowledge Distillation

Wilson Lau[1], Laura Aaltonen, MD PhD[2], Martin Gunn, MB ChB[2], Meliha Yetisgen, PhD[1,3]
[1]Department of Biomedical and Health Informatics, [2]Department of Radiology,
[3]Department of Linguistics, University of Washington, Seattle, WA

**Abstract**

*Selecting radiology examination protocol is a repetitive, and time-consuming process. In this paper, we present a deep learning approach to automatically assign protocols to computed tomography examinations, by pre-training a domain-specific BERT model ($BERT_{rad}$). To handle the high data imbalance across exam protocols, we used a knowledge distillation approach that up-sampled the minority classes through data augmentation. We compared classification performance of the described approach with n-gram models using Support Vector Machine (SVM), Gradient Boosting Machine (GBM), and Random Forest (RF) classifiers, as well as the $BERT_{base}$ model. SVM, GBM and RF achieved macro-averaged F1 scores of 0.45, 0.45, and 0.6 while $BERT_{base}$ and $BERT_{rad}$ achieved 0.61 and 0.63. Knowledge distillation boosted performance on the minority classes and achieved an F1 score of 0.66.*

## Introduction

When an advanced imaging order is placed, e.g. for computed tomography (CT) or magnetic resonance imaging (MRI), a radiologist often has to select the most applicable imaging protocol for the imaging technologist to perform the examination. The imaging protocol contains the technical parameters for CT or MRI image acquisition. For CTs and MRIs, this includes the number of sequences, use of intravenous or oral contrast, scanning planes and other technical parameters to ensure that the image acquisition is best suited to answer the clinical question being asked by the ordering provider. The radiologist gives the protocol decision based on the free-text clinical information in the electronic order and the suggested examination by the ordering physician. This manual protocoling process can be time-consuming, may delay performing timely imaging, and result in unnecessary variability in the techniques used for image acqusition[1].

Radiologists are one of the highest paid medical specialties, with an average salary of USD $485,000 in 2020, according to the 2020 physician compensation report published by Doximity. Radiologists' primary clinical and revenue generating task is image interpretation, but their time is often spent performing 'non-interpretive tasks', which are non-revenue generating and often can distract them from performing quality image interpretations. Radiologists may spend 37-44% of their typical workday performing non-interpretive tasks, and 3.5-6.2% of their time protocoling[2,3]. For clinically urgent studies, radiologists may be interrupted to protocol non-urgent examinations, which reduces productivity, potentially impairs interpretive accuracy and lengthens report turnaround time (a radiology service quality metric)[4,5].

Radiology exam protocoling is a repetitive and fairly simple classification task. It is therefore a strong candidate for automation. By applying machine learning techniques to protocoling, radiologists could spend a greater proportion of their time performing interpretive tasks, thereby improving the cost-effectiveness of a radiology practice, reducing interruptions for protocoling, improve interpretation accuracy and shorten report turnaround time. Integration of a ML protocoling pipeline into the protocoling software that radiologists use (usually the radiology information system) would permit radiologists to obviate the need to protocol examinations altogether or, in some cases, reduce the number of protocoling steps by suggesting the correct protocol or simply flagging more complex protocols for a radiologist to 'manually' protocol.

In this paper, we defined the task of protocol assignment as a classification task. We used structured radiology exam meta-data (exam name and code provided by referring physician) and patient demographics (age and gender) as well as free text diagnoses and history information to automatically assign a radiology protocol. Table 1 presents an example of the radiology examination data from our dataset. Information listed in Table 1 is available to radiologists when they assign protocols manually. In our experiments, we (1) compared different statistical ML models to the state-of-the-art BERT[6] model for radiology protocol classification task, (2) evaluated the BERT model pre-trained on general domain ($BERT_{base}$) in comparison to a BERT model pre-trained on our radiology corpus ($BERT_{rad}$), and (3) applied deep learning knowledge distillation approach to tackle high data imbalance in our dataset.

| Exam metadata | | Demographics | | Patient history | | Protocol |
|---|---|---|---|---|---|---|
| Code | Name | Sex | Age | History | Diagnosis | |
| CABDWC | CT ABDOMEN W CONTRAST | 2 | 67 | heart failure, hepatic vein | concern for liver laceration post procedure, post biopsy, on apixaban | BODY CT Liver 2 phase for hypervascular liver metastases (art venous, no delay) |

**Table 1.** Example examination data from our dataset.

### Related Work

Automating radiology protocol selection has been investigated in previous studies[7–9]. Brown et al. compared three different ML models, including support vector machine (SVM), gradient boosting machine (GBM), and random forest (RF), to classify MRI protocol selection[7]. They used bag-of-words approach with unigrams to represent features for the text data and combined them with the structured variables (age, sex, location and ordering service). The dataset consisted of 7487 observations. Since each protocol can consist of a sequence of procedures, it is considered a multi-label classification task. They trained 41 binary classifiers for each model to predict each procedure in a sequence. The three ML algorithms included in this study demonstrated similar performance. GBM achieved 86% precision and 80% recall. SVM achieved 83% precision and 82% recall, followed by RF with 85% precision and 80% recall. In another study[8], the same authors employed similar ML approaches to protocol and prioritize MRI brain examinations. Their best classifier using RF achieved 82% precision and 83% recall. In this paper, we used SVM, GBM, and RF as baselines to compare the performance of our proposed classification approach.

Trivedi et al. used IBM Watson to determine the use of intravenous contrast for musculoskeletal MRI protocols by analyzing only clinical texts[10]. The task was to classify a free-text clinical indication into one of the two labels "with contrast" or "non-contrast". The dataset consisted of 650 positive and 870 negative labels. Watson achieved over 90% precision and 74% recall. The overall performance is similar to their ensemble model comprising 8 traditional statistical models (SVM, scaled linear discriminant analysis, boosting, bagging, classification and regression tree, RF, Lasso and elastic-net regularized generalized linear model, maximum entropy). Although they claimed that Watson's classifier was based on deep learning, no specific details about the model architecture and hyperparameters were provided by IBM.

One research conducted by Kalra et al. is the most similar to our study. They developed two statistical ML models and one deep learning model to automate CT and MRI protocol assignment. The dataset contained 18000 CT and MRI examinations in 108 unique protocols. Similar to our dataset, their protocol frequency distribution is highly imbalanced with the 5 most commonly assigned protocols making up 49% of the entire dataset. They trained a k-nearest neighbor and a random forest classifier using TF-IDF feature vectors on unigrams from clinical texts. Interestingly, they excluded structured data elements such as age and gender, which could be strong predictor variables. The performance results from the top two classifiers, RF (80% precision, 82% recall) and DNN (82% precision, 84% recall), were comparable. However, they only reported weighted micro-averages and did not report performance metrics per protocol. Hence, we do not know how the model performed on the minority classes.

Our main contribution in this paper is the feasibility analysis of applying transfer learning using pre-trained language models for protocol classification task. In our experiments, we first evaluated the three ML approaches (SVM, GBM, and RF) presented in previous studies, and observed similar micro-averaged classification performance. Compared to other published datasets, our dataset is relatively larger in which 57% of the data fall into two specific protocol groups. To handle such high data imbalance, we presented a knowledge distillation approach by up sampling the minority classes through data augmentation. We presented the classification result for each protocol group and showed the performance gain for the minority classes.

### Methods

**Dataset:**

Our dataset included 35085 radiology body CT examinations performed at 7 hospital-based and clinic-based imaging sites between January 2018 and June 2019. The data was extracted from the University of Washington radiology information system. As demonstrated in Table 1, each exam is represented with 4 structured data fields including exam meta-data (exam code, protocol code) and patient demographics (age, gender) as well as 2 unstructured fields to capture patient history (history, diagnosis). Table 2 describes the word level statistics on the two unstructured fields. In our initial analysis of the dataset, we observed that the lengths of the unstructured data are relatively short (average numbers of words for history and diagnosis fields were 8 and 10 with standard deviations 6.57 and 8.6 respectively). 4759 (13.6%) examinations contained no history data and 3 (0.01%) examinations contained no diagnosis data.

|  | Min | Max | Mean | Median | Standard deviation |
|---|---|---|---|---|---|
| **History** | 0 | 47 | 8 | 6 | 6.57 |
| **Diagnosis** | 0 | 108 | 10 | 8 | 8.6 |

**Table 2.** Word statistics on unstructured fields.

In addition, we observed that some of the same protocols had multiple protocol codes, reflecting different codes for the identical protocols performed at different imaging sites. To remove this inconsistency, we manually categorized the protocol codes into 27 unique "protocol groups"; each group unified identical protocols with different codes. We excluded 2 groups that had less than 20 examinations in our experiments (*CT CA Oral Only* and *CT Abdomen IV Only*). Table 3 shows the examination frequency with percentages for each protocol group. As can be observed from Table 3, the dataset is highly imbalanced, with the first two protocol groups constituting 57% of the entire dataset. The distribution of examination frequency among the groups has a mean of 1299, median of 200 and standard deviation of 2706.

|  | Protocol group | Fre. | % |  | Protocol group | Fre. | % |
|---|---|---|---|---|---|---|---|
| 1. | CT CAP IV and Oral | 11911 | 33.95% | 15. | CT Pancreas Mass 3 Phase | 202 | 0.58% |
| 2. | CT Abdomen Pelvis w IV Only | 8057 | 22.96% | 16. | CT Abdomen No Contrast | 195 | 0.56% |
| 3. | CT CAP IV Only | 3351 | 9.55% | 17. | CT CA IV and Oral | 194 | 0.55% |
| 4. | CT Abdomen Pelvis w IV and Oral | 2941 | 8.38% | 18. | CT Pelvis IV Only | 192 | 0.55% |
| 5. | CT Renal Mass | 2036 | 5.80% | 19. | CT Abdomen IV and Oral | 173 | 0.49% |
| 6. | CT Liver 3 Phase | 1652 | 4.71% | 20. | CT Pancreas Mass 2 Phase | 143 | 0.41% |
| 7. | CT Abdomen Pelvis No Contrast | 931 | 2.65% | 21. | CT Abdomen Pelvis w Oral only | 132 | 0.38% |
| 8. | CT IVP 50 yrs + | 854 | 2.43% | 22. | CT CA No Contrast | 75 | 0.21% |
| 9. | CT CAP Oral Only | 531 | 1.51% | 23. | CT Pelvis Cystogram | 68 | 0.19% |
| 10. | CT CAP No Contrast | 336 | 0.96% | 24. | CT Liver 2 Phase | 51 | 0.15% |
| 11. | CT Abd Pel Enterography | 297 | 0.85% | 25. | CT Pelvis IV and Oral | 42 | 0.12% |
| 12. | CT Liver 4 Phase | 252 | 0.72% | 26. | CT CA Oral Only (excluded) | 15 | 0.04% |
| 13. | CT CA IV Only | 226 | 0.64% | 27. | CT Abdomen IV Only (excluded) | 8 | 0.02% |
| 14. | CT IVP < 50 | 220 | 0.63% |  |  |  |  |

**Table 3.** Distribution of examinations across protocols.

**Approach:**

We trained a deep learning classifier using the state-of-the-art neural language model, BERT[6] to automatically assign protocols to computer tomography (CT) examinations. Specifically, we fine-tuned the Google pre-trained model $BERT_{base}$ with a linear layer on top using cross-entropy loss. We formulated the task as a single-sequence classification task by first transforming the structured and unstructured data into the following template: "*Exam is <exam code>. Sex is <gender>. Age at Exam <age>. History: <history>. Diagnosis: <diagnosis>*" and subsequently classifying it into one of 25 protocol groups listed in Table 3. We observed that the mean and median of number of characters in the templated data are 192 and 178. In order to capture context presented in the training instances, we set the maximum sequence length parameter of the BERT model to be 200 with a batch size of 48. We followed the suggestions described in the BERT paper and used the Adam optimizer with a learning rate of 2-e5. We fine-tuned the BERT model for 4 epochs.

Conceptually, BERT learns the relations between words by randomly masking words in a sequence with a [MASK] token and then trains itself to predict them from the context of the unmasked ones. Additionally, it learns the sentence relationships by training itself to predict if the second sentence in a pair is truly following the first sentence in the corpus. These two learning tasks allow BERT to self-train and capture the context of language used in an unlabeled corpus before transferring all parameters to down-stream applications. Previous studies showed promising results of using BERT in clinical applications. Examples include chest x-ray reports classification[11], and relation extraction in clinical and biomedical domain[12,13]. Since $BERT_{base}$ was originally pre-trained on BookCorpus and English Wikipedia, to fully encode the semantic context in clinical and biomedical text, it has been shown that further training $BERT_{base}$ on MIMIC and PubMed data can boost the performance of named entity recognition in the biomedical domain[14,15]. Inspired by these studies, we further pre-trained $BERT_{base}$ on our radiology protocol corpus and named it $BERT_{rad}$. We repeated the same experiment with $BERT_{rad}$ using the same hyperparameters listed above. All BERT experiments were implemented with Huggingface's transformer library[16].

Knowledge distillation:

Imbalanced class distribution usually leads to poor classification results on the minority classes[17]. When dealing with imbalanced datasets, a popular approach is to use the Synthetic Minority Oversample Technique (SMOTE)[18] which generates new artificial samples for the minority classes by interpolating the nearest neighbors of the existing samples. This method reduces the likelihood of overfitting minority classes commonly observed in random over sampling approach. However, because the inputs of the BERT model include positional embeddings and WordPiece tokenization with special classifier token [CLS] and separator token [SEP], synthesizing these input values in vector space using interpolation will lose the context of the tokens in the samples.

Recent studies have successfully demonstrated the possibility to transfer task specific knowledge from the large BERT model to a smaller neural architecture without significant performance degradation[19–21], using a technique called knowledge distillation. The process involves training a second model (student) to match the predictions from the first model (teacher). We hypothesized that by transferring knowledge specific to the minority classes from the $BERT_{rad}$ model to a second BERT model, we could improve the classification performance on the minority classes. In particular, we aimed to train a student model that could outperform the teacher with identical neural architecture. Furlanello et al. referred to this approach as Born-Again Neural Network (BAN)[22], which has been shown to produce better results in both single and multi-task settings[23]. During the knowledge distillation process, the raw predictions from the teacher model, known as logits, are being used as "soft labels" for training the student model. As Hinton et al. suggest, the distribution in the logits, even among incorrect predictions, contains information about how the teacher model is generalizing, thereby offering more training signals than one-hot categorical labels[19].

To effectively transfer knowledge about the minority classes to a student model, a large unlabeled dataset is needed to generate enough soft labels from the teacher model. In this study, we applied Tang et al.'s data augmentation techniques to synthesize masked data in order to allow the teacher to fully express its knowledge[21]. To augment a given training instance, we randomly sampled a number $P$ from the uniform distribution [0,1]. If $P < 0.1$, we randomly replaced a word in the history and diagnosis section with the [MASK] token. If $P$ is between 0.1 and 0.2, we randomly replaced a word with another word in the training set that has the same POS tag. Finally, we randomly replaced an n-gram (n ∈ [1,3]) in the training instance with the [MASK] token. This technique is similar to the masking procedure employed in BERT's masked language model. We repeated this augmentation process to generate 30 new instances, without duplication, for each training instance. We evaluated different numbers of augmented instances (25, 30, 35, 40, 50) by running 5-fold cross validation with the augmented data. Our evaluation showed that the experiment with 30 augmented instances achieved the best result. In addition, we wanted to limit the augmented sample size of the dominant classes and therefore set a maximum sampling limit of 12000, such that the final sample size of each class after augmentation could not exceed 12000. We then ran inferencing on the augmented dataset using the teacher model $BERT_{rad}$ to generate soft labels for distillation. Finally, we initialized a student $BERT_{rad}$ model with a different random seed and trained it to imitate the teacher by minimizing the mean squared error (MSE) between the student's logits and teacher's logits. At the same time, we allowed the student model to surpass the teacher by training with the true labels by minimizing the cross-entropy loss against the one-hot multi-class labels:

$$L_{distill} = \alpha * L_{cross-entropy} + (1 - \alpha) * L_{MSE}$$

where $\alpha$ is the ratio of true labels within a single batch of training samples. After each iteration of knowledge distillation, the student model became the teacher for next generation.

Baselines: For our prediction task, we trained three separate ML models with Support Vector Machine (SVM), Gradient Boosting Machine (GBM) and Random Forest (RF) as baselines and compared their performance against our proposed deep learning classifier. To train the baselines, we transformed the unigrams and bigrams of the history and diagnosis notes into feature space using TF-IDF before combining with the numeric values in the structured data. The baselines were trained with the same features. All statistical classification modelling was implemented using the Scikit-learn machine learning python package [24].

## Results

We used 5-fold cross validation to evaluate the general performance of the models. For each fold, the models were trained on the same training data and evaluated on the same held-out test data. We used precision, recall, and F1-score as metrics to measure the performance. The overall macro-average and weighted micro-average results are presented in Table 4. To generate macro-averaged results, we first calculated the metrics for each class and calculated the average of them giving each class equal weight. To generate weighted micro-averaged metrics, we calculated the metric averages weighted by the number of true labels in each class. As can be observed, the micro-average results are largely similar due to the bias towards the majority classes. In the macro-average results, among the baselines, RF performed the best with 0.60 F1-Score. Both SVM and GBM produced 0.45 F1-score. The SVM in general produced higher precision and lower recall, when compared to GBM. The classifiers based on BERT models performed better than the SVM, GBM and RF baselines. Furthermore, the in-domain $BERT_{rad}$ produced 0.2 higher macro F1 score than the out-of-domain $BERT_{base}$ model (0.63 versus 0.61).

To mitigate the high data imbalance, two resampling experiments were performed with the $BERT_{rad}$ model. The resampling was performed only on the training data while the validation data were kept the same. First, we under-sampled the 2 majority classes by randomly removing some training instances such that their sample sizes matched the size of the third largest protocol group (#3). The result shows that the macro-average F1 dropped 0.24 and the weighted micro-average F1 dropped 0.2 due to the misclassification of the majority classes given their smaller sample sizes. In the over-sampling experiment, the training instances in the minority classes were randomly replicated so that their sample sizes matched the size of the second largest protocol group (#2). The result shows no performance improvement in the macro-average F1 but degradation in the weighted micro-average. This can be caused by overfitting the large number of duplicate training samples in the minority classes[18]. The results also show that the BAN models{2,3} achieved better macro-average performance than $BERT_{base}$ and $BERT_{rad}$, without any degradation in micro-average performance. More specifically, the macro-average F1 in generations of student models (BAN{1,2,3}) improved, suggesting that the classifiers achieved better performance in predicting the minority classes through knowledge distillation. We also observed that the performance saturated after training the second generation of BAN student model. This finding is similar to the one reported by Furlanello et al[22].

| Model | Macro average | | | Micro (Weighted) average | | |
|---|---|---|---|---|---|---|
| | **Precision** | **Recall** | **F1** | **Precision** | **Recall** | **F1** |
| SVM | 0.60 | 0.42 | 0.45 | 0.79 | 0.80 | 0.79 |
| GBM | 0.46 | 0.46 | 0.45 | 0.80 | 0.81 | 0.80 |
| RF | 0.63 | 0.59 | 0.60 | 0.83 | 0.83 | 0.83 |
| $BERT_{base}$ | 0.68 | 0.60 | 0.61 | 0.84 | 0.84 | **0.84** |
| $BERT_{rad}$ | 0.67 | 0.62 | 0.63 | 0.84 | 0.84 | **0.84** |
| $BERT_{rad}$ undersample | 0.42 | 0.38 | 0.39 | 0.63 | 0.66 | 0.64 |
| $BERT_{rad}$ oversample | 0.63 | 0.63 | 0.63 | 0.83 | 0.82 | 0.82 |
| BAN1 | 0.68 | 0.64 | 0.65 | 0.84 | 0.84 | **0.84** |
| BAN2 | 0.69 | 0.65 | **0.66** | 0.84 | 0.84 | **0.84** |
| BAN3 | 0.69 | 0.65 | **0.66** | 0.84 | 0.84 | **0.84** |

**Table 4.** Comparison of model results. BAN{1,2,3} denotes the 1st, 2nd and 3rd generation of knowledge distillation.

Table 6 presents performance results at the protocol level. Our results showed that the BERT models generally outperformed statistical baselines among the protocol groups. From the protocol group #2 to #9, the difference in F1 scores between the best performing statistical model and the best performing BERT model is between 1% to 5%. In some specific groups, such as "CT CAP No Contrast" (#10) and "CT Abd Pel Enterography" (#11), we observed much larger improvement. Overall, we did not observe a substantial improvement in $BERT_{rad}$ compared to $BERT_{base}$. The small size of the dataset and sparseness of the unstructured data fields resulted this outcome. Nonetheless, $BERT_{rad}$ was able to outperform $BERT_{base}$ in some protocol groups by capturing the context of words that are not common in general domain. For example, in the protocol group "CT Abd Pel Enterography" (#11), the word *hernia*, which describes the symptom that a tissue pushes through the abdominal opening, appeared in over 79% of the diagnosis fields, while another word *CREATININE*, a compound that indicates the level of kidney function, appeared in over 73% of the history fields. These two medical terms are not commonly seen in the general corpora. By pre-training on the radiology corpus, $BERT_{rad}$ was able to learn better contextual representation of these medical terms and outperformed $BERT_{base}$ by 0.07 F1 in that protocol group. We observed similar improvement in protocol groups "CT Abdomen IV and Oral" (#19) and "CT Abdomen Pelvis w Oral only" (#21).

One interesting observation was $BERT_{base}$ model's substantially low F1-score of 0.16 for group "CT IVP < 50" (#14) when compared to the F1-scores (SVM: 0.61, GBM: 0.73, RF: 0.88) of statistical baselines. Further investigation showed that 87% of the false negatives for "CT IVP < 50" (#14) were misclassified to "CT IVP 50 yrs +" (#8) by $BERT_{base}$. Because the main difference between these two protocol groups is age of patient, the age feature by itself offered high information gain to allow RF to learn a more robust model. On the other hand, deep learning models require considerably large volume of data to extract patterns from high-dimensional data points. The smaller sample size of protocol group #14 limited the $BERT_{base}$ model to learn to differentiate from protocol group #8. However, data augmentation in the knowledge distillation process eventually supplied additional training signals for the model to generalize, leading to the similar performance levels as RF.

Although knowledge distillation enabled the BERT models to improve overall performance on the minority classes, one particular protocol group that was not correctly classified by any models was "CT Liver 2 Phase" (#24). Our error analysis showed that the models misclassified some #24 cases to "CT Liver 3 Phase" (#6) because of similar patient diagnosis and history. Table 5 presents one of these cases.

| Protocol group | History | Diagnosis |
| --- | --- | --- |
| 6. CT Liver 3 Phase | Last creatine level:CREATININE 0.92 | ABDOMEN W/CONTRAST; 6MO REPEAT F/U FOR HCC SURVEILLANCE, S/P LIVER TRANSPLANT |
| 24. CT Liver 2 Phase | Last creatine level:CREATININE 0.81 | ABDOMEN W/CONTRAST; TO EVALUATE SIZE OF PSEUDOCYST, S/P LIVER TRANSPLANT |

**Table 5.** Examinations in two different protocol groups with similar history and diagnosis.

While these were the correct protocol assignments in clinical practice, because #24 only constituted 0.15% of the training data and was 30 times less than #6, there were not enough data to train the models to differentiate #24 from #6. Additionally, we found that some #24 cases were misclassified to "CT CAP IV and Oral" (#1) because of the exact same history and diagnosis found in #24. For instance, there were 6 cases with history of "*ORAL and IV Contrast*" and diagnosis of "*2 Phase Liver, PNET Metastatic*" assigned to protocol #1 and 1 case with the same history and diagnosis assigned to #24. Without any additional clinical information to help differentiate the two protocol assignments, the models simply inferred to the group that was more dominant in the training data.

| Protocol group | Exam count | SVM | GBM | RF | BERT$_{base}$ | BERT$_{rad}$ | BAN1 | BAN2 | BAN3 |
|---|---|---|---|---|---|---|---|---|---|
| 1. CT CAP IV and Oral | 2382 | 0.92 | 0.91 | **0.93** | **0.93** | **0.93** | **0.93** | **0.93** | **0.93** |
| 2. CT Abdomen Pelvis w IV Only | 1612 | 0.85 | 0.86 | 0.87 | **0.88** | 0.87 | 0.87 | 0.87 | 0.87 |
| 3. CT CAP IV Only | 670 | 0.78 | 0.73 | 0.78 | **0.80** | **0.80** | 0.79 | 0.79 | **0.80** |
| 4. CT Abdomen Pelvis w IV and Oral | 588 | 0.66 | 0.66 | 0.66 | **0.70** | 0.69 | 0.67 | 0.67 | 0.67 |
| 5. CT Renal Mass | 407 | 0.83 | 0.89 | 0.91 | 0.92 | **0.93** | 0.92 | **0.93** | 0.92 |
| 6. CT Liver 3 Phase | 330 | 0.83 | 0.82 | 0.85 | **0.87** | **0.87** | 0.86 | 0.86 | **0.87** |
| 7. CT Abdomen Pelvis No Contrast | 186 | 0.42 | 0.73 | 0.75 | **0.77** | 0.76 | **0.77** | **0.77** | **0.77** |
| 8. CT IVP 50 yrs + | 171 | 0.81 | 0.90 | 0.92 | 0.84 | 0.84 | 0.91 | **0.93** | 0.92 |
| 9. CT CAP Oral Only | 107 | 0.29 | 0.56 | 0.31 | 0.58 | 0.59 | 0.59 | **0.61** | **0.61** |
| 10. CT CAP No Contrast | 67 | 0.01 | 0.01 | 0.07 | **0.16** | **0.16** | 0.14 | 0.13 | **0.16** |
| 11. CT Abd Pel Enterography | 59 | 0.54 | 0.41 | 0.50 | 0.53 | 0.60 | **0.63** | 0.61 | 0.62 |
| 12. CT Liver 4 Phase | 51 | 0.07 | 0.33 | 0.58 | 0.66 | **0.69** | 0.68 | 0.67 | 0.67 |
| 13. CT CA IV Only | 45 | 0.59 | 0.49 | 0.75 | **0.78** | **0.78** | 0.77 | **0.78** | **0.78** |
| 14. CT IVP < 50 | 44 | 0.61 | 0.73 | **0.88** | 0.16 | 0.24 | 0.78 | **0.88** | 0.87 |
| 15. CT Pancreas Mass 3 Phase | 41 | 0.46 | 0.36 | 0.58 | 0.64 | 0.62 | 0.63 | **0.67** | 0.65 |
| 16. CT Abdomen No Contrast | 39 | 0.61 | 0.43 | 0.79 | **0.82** | 0.81 | 0.81 | **0.82** | 0.80 |
| 17. CT CA IV and Oral | 39 | 0.12 | 0.10 | 0.56 | 0.61 | **0.62** | 0.61 | 0.61 | 0.59 |
| 18. CT Pelvis IV Only | 39 | 0.78 | 0.77 | 0.80 | **0.83** | **0.83** | 0.82 | **0.83** | **0.83** |
| 19. CT Abdomen IV and Oral | 35 | 0.06 | 0.03 | 0.36 | 0.41 | 0.45 | **0.48** | **0.48** | **0.48** |
| 20. CT Pancreas Mass 2 Phase | 28 | 0.05 | 0.15 | 0.21 | **0.29** | 0.28 | 0.28 | **0.29** | **0.29** |
| 21. CT Abdomen Pelvis w Oral only | 26 | 0.00 | 0.05 | 0.26 | 0.23 | 0.37 | **0.39** | 0.38 | 0.37 |
| 22. CT CA No Contrast | 15 | 0.19 | 0.09 | 0.71 | **0.76** | 0.75 | 0.74 | 0.75 | **0.76** |
| 23. CT Pelvis Cystogram | 14 | 0.69 | 0.27 | 0.95 | 0.95 | 0.95 | 0.95 | 0.95 | **0.96** |
| 24. CT Liver 2 Phase | 10 | 0.00 | **0.04** | 0.00 | 0.00 | 0.00 | 0.00 | 0.00 | 0.00 |
| 25. CT Pelvis IV and Oral | 8 | 0.00 | 0.03 | 0.03 | 0.15 | **0.31** | 0.28 | 0.25 | 0.28 |

**Table 6.** Comparison of model results in F1 for each protocol group on the test set.

## Conclusion

In this study, we presented a novel ML approach using pre-trained language models to help radiologists automatically assign protocols based on patient demographic and history data. The results showed that overall pre-trained language models performed better than traditional n-gram models. Additionally, we demonstrated that knowledge distillation improved overall classification performance for the majority of under-represented groups. Since many real-world biomedical datasets are intrinsically imbalanced (e.g. the prevalence of certain uncommon cancer types or chronic diseases[25,26] ), we think that this technique could be useful in many classification problems involving clinical text using pre-trained language models. In future studies, we plan to integrate the machine learning models into actual radiologist workflow and compare the agreement between radiologists and models. The dataset and experimentation source codes will be shared with the research community[*].

---

[*] https://github.com/wilsonlau-uw/Automatic-Assignment-of-Radiology-Examination

## Acknowledgements

This publication was partially supported by the National Center For Advancing Translational Sciences of the National Institutes of Health under Award Number UL1 TR002319. Research and results reported in this publication was facilitated by the generous contribution of computational resources from the University of Washington Department of Radiology.